%% file: eacl2021.tex
\documentclass[11pt,a4paper]{article}
\usepackage[hyperref]{eacl2021}
\usepackage{times}
\usepackage{latexsym}
\usepackage{graphicx}
\usepackage{amsmath}
\usepackage{amssymb}
\usepackage{textcomp}
\usepackage{bbm}
\usepackage{multirow}
\usepackage{tabularx}
\usepackage{makecell}
\usepackage{enumitem}
\usepackage{ragged2e}
\usepackage{booktabs}
\usepackage{subfig}
\usepackage{tablefootnote}
\usepackage{footnotehyper}

\usepackage{microtype}

\aclfinalcopy % Uncomment this line for the final submission
\def\aclpaperid{956} %  Enter the acl Paper ID here

\title{Hidden Biases in Unreliable News Detection Datasets}

\author{Xiang Zhou$^1$, Heba Elfardy$^2$, Christos Christodoulopoulos$^2$\\
   \textbf{Thomas Butler$^2$, Mohit Bansal$^1$} \\
  $^1$UNC Chapel Hill \quad $^2$Amazon\\
  \texttt{\{xzh, mbansal\}@cs.unc.edu} \\ 
  \texttt{\{helfardy, tombutl\}@amazon.com, chrchrs@amazon.co.uk}}

\date{}

\begin{document}
\maketitle

\input{0abstract.tex}
\input{1introduction.tex}

\input{2related}
\input{3collection}
\input{4split}
\input{6conclusion}

\section*{Acknowledgments}
We thank the reviewers for their helpful comments. 
XZ interned at Amazon.
This work was also supported by ONR Grant N00014-18-1-2871, DARPA YFA17-D17AP00022, and DARPA KAIROS Grant FA8750-19-2-1004. The views contained in this article are those of the authors and not of the funding agency.

\bibliography{eacl2021}
\bibliographystyle{acl_natbib}

\end{document}

%% file: 0abstract.tex
\begin{abstract}
    Automatic unreliable news detection is a research problem with great potential impact. 
    Recently, several papers have shown promising results on large-scale news datasets with models that only use the article itself without resorting to any fact-checking mechanism or retrieving any supporting evidence. 
    In this work, we take a closer look at these datasets. While they all provide valuable resources for future research, we observe a number of problems that may lead to results that do not generalize in more realistic settings. Specifically, we show that selection bias during data collection leads to undesired artifacts in the datasets. 
    In addition, while most systems train and predict at the level of individual articles, overlapping article sources in the training and evaluation data can provide a strong confounding factor that models can exploit. In the presence of this confounding factor, the models can achieve good performance by directly memorizing the site-label mapping instead of modeling the real task of unreliable news detection. We observed a significant drop ($>$10\%) in accuracy for all models tested in a clean split with no train/test source overlap. 
    Using the observations and experimental results, we provide practical suggestions on how to create more reliable datasets for the unreliable news detection task. We suggest future dataset creation include a simple model as a difficulty/bias probe and future model development use a clean non-overlapping site and date split.\footnote{
Our code is publicly available at \url{https://owenzx.github.io/unreliable_news}}
\end{abstract}

%% file: 1introduction.tex
\begin{table*}[t]
\centering
% \small
\renewcommand\theadfont{\bfseries}

\AtBeginEnvironment{tabularx}{\setlist[enumerate, 1]{wide, leftmargin=*, itemsep=0pt, before=\vspace{-\dimexpr\baselineskip +2 \partopsep}, after=\vspace{-\baselineskip}}}
\begin{small}
 \begin{tabularx}{\linewidth}{XXX}
        \toprule %
        \thead{Data Collection} & \thead{Dataset Construction} & \thead{Experiment Design} \\%
        \cmidrule[\lightrulewidth](lr){1-3}
        \begin{enumerate}
            \item Collect from less biased or unbiased resources (e.g. original news outlets). (Sec. \ref{sec:biases})
            \item Collect from diverse resources (in terms of sources, topics, time, etc.). (Sec. \ref{sec:biases}, \ref{sec:split})
            \item Collect precise article-level labels if possible. (Sec. \ref{sec:collect})
        \end{enumerate} & \begin{enumerate}%
            \item Examine the most salient words to check for biases in the datasets. (Sec. \ref{sec:biases})
            \item Run simple BoW baselines to check how severe the bias is. (Sec. \ref{sec:split})
            \item Provide train/dev/test splits with non-overlapping source/time. (Sec.~\ref{sec:sp_source}, \ref{sec:sp_time})
        \end{enumerate} & \begin{enumerate}%
            \item Apply debasing techniques when developing models on biased datasets. (Sec. \ref{sec:biases})
            \item Check the performance on sources/dates not in your training set. (Sec. \ref{sec:sp_source}, Sec. \ref{sec:sp_time})
            \item Check the performance on sources with limited examples. (Sec. \ref{sec:error})
            \item Test your model on multiple complementary datasets (e.g. with different domains, styles, etc.). (Sec. \ref{sec:biases}, \ref{sec:error})
        \end{enumerate}\\%
        \bottomrule
    \end{tabularx}
\end{small}
% \vspace{-4pt}
\caption{Suggestions for data collection, dataset construction and experiment design for unreliable news research.}
% \vspace{-9pt}
\label{tab:checklist}
\end{table*}

\section{Introduction}

The proliferation of unreliable news is widely acknowledged~\cite{del2016spreading,lazer2018science,vosoughi2018spread}, and its identification is a socially important problem. 
In this work we use the label unreliable news as a broad term for all unverifiable and misleading news content, regardless of whether the content is malicious (targeted misinformation) or not. Accordingly, while specific definitions vary in different datasets used in this work, we refrain from using the term ``fake'' since identifying the intent of the author(s) is beyond the scope of this work. 
To mitigate the problem of surfacing unreliable news content, various websites (e.g., PolitiFact\footnote{\url{https://www.politifact.com/}}, 
Media Bias/Fact Check (MBFC)\footnote{\url{https://mediabiasfactcheck.com/}}, GossipCop\footnote{\url{https://www.gossipcop.com/}}, etc.) determine the reliability of news by manually fact-checking the important claims in given news articles.
Beyond requiring investigative expertise, manual fact-checking is time-consuming and is thus limited to only a small set of selected news articles.

Recent research has explored automating this process using machine learning methods to automatically determine news veracity \cite{perez-rosas-etal-2018-automatic,baly2018predicting,nie2019combining,wright2020fact}. These efforts were made possible due to the availability of large-scale unreliable news detection datasets \cite{horne2018sampling,shu2017fake,wang2017liar}. 
In our work, we examine if these datasets accurately reflect the real difficulty of this task or if there are any hidden biases in the datasets. Specifically, we study different methods of dataset construction (e.g., how the data was collected, how the data was split, etc.) and show that the assessed difficulty of the task is sensitive to how carefully different factors are considered when building and using these datasets.

Our investigation begins with data collection procedures: we look at the source of news stories (news outlets, social media, fact-checking websites, etc.) as well as the annotation process (number of labels, granularity of labels, article- or site-level annotation). 
We discuss the pros and cons of each approach and point out some hidden pitfalls. 
Using FakeNewsNet~\cite{shu2017fake} as an example, we demonstrate how selection biases in data collection can lead to undesired biases in the created datasets.

Moving beyond data collection, we examine two commonly applied ways of splitting the dataset for training and testing that help the model achieve high performance without correctly modeling the task.
Specifically, we show that using a disjoint set of sites/news outlets for training and test data significantly decreases the models' performance~($>$10\%) and that the drop in performance is related to how similar (or dissimilar) the sites in both sets are (reflected by various site-level distributional distance metrics including L2, cos, EMD, etc.). 
Additionally, we also examine the effect of time overlap between both train and test sets. We observe that different news outlets are likely to have similar content in a small time window (i.e., the same story gets covered by multiple outlets within a day or a few days period). While we do not find any evidence that the studied models exploit this factor, we nevertheless suggest that future datasets are split both by time and site/news outlet.

In summary, our main contributions are: (1) showing how data collection procedures can lead to systematic biases in unreliable news datasets, (2) demonstrating how confounding factors--—such as site/news outlet and time—--in these datasets can degrade their quality and lead to underestimating the difficulty of the task, and finally (3) suggesting possible mechanisms to avoid these biases and confounding factors when building new datasets. 
To facilitate future research, we also provide a list of practical suggestions for  data collection, dataset construction, and experiment design in Table~\ref{tab:checklist}.

%% file: 2related.tex
\begin{savenotes}
\begin{table*}[t]
\centering
\small
\begin{tabular}{lccc}
\toprule 
Dataset & Size    & Article Source & Label Type      \\
\midrule 
NELA\footnotemark{}      & 136K/713K/1.12M   & News outlets & Site-level            \\
% \midrule
FakeNewsNet\footnotemark{} & 603K   & Fact-checking websites & Article-level   \\
% \midrule
r/Fakeddit\footnotemark{} & 1.06M & Social Media (Reddit) & Site-level \\                           
\bottomrule 
\end{tabular}

\caption{Statistics and properties of three recent large-scale unreliable news datasets. The three statistics of NELA dataset sizes correspond to its three versions released in 2017, 2018 and 2019, respectively.}
\label{tab:datasets}
\end{table*}
\end{savenotes}

\section{Related Work}

\paragraph{Unreliable News Detection.}
Unreliable news detection and other news veracity related tasks have been receiving an increasing focus as news sources have become more accessible in recent years. A lot of effort has been put into collecting high-quality datasets.
\citet{wang2017liar,shu2017fake} collected manually labeled statements or news articles from fact-checking websites. The NELA datasets~\cite{horne2018sampling,norregaard2019nela,gruppi2020nela} scrape news articles directly from news outlets and use the manually annotated labels from Media Bias/Fact Check (MBFC) as site-level annotations. Social media is also a popular resource for collecting news stories~\cite{nakamura2020fakeddit,santia2018buzzface, mitra2015credbank}. Researchers have also collected datasets for various related topics, such as rumor detection~\cite{kwon2017rumor,ma2016detecting}, and propaganda detection~\cite{da2020survey,barron2019proppy}. Besides classifying the veracity of news articles, researchers have also explored related problems, such as predicting the reliability of news sites~\cite{baly2018predicting}, 
 identifying fact-check worthy sentences~\cite{wright2020fact}, among other tasks. Several recent papers also focus on measuring the trustworthiness of single statements~\cite{wang2017liar,pomerleau2017fake, alhindi-etal-2018-evidence}. In this work, we focus on article-level classification because of its relevance to applications, like news feeds, that operate at the article level. 
\paragraph{Pitfalls in Data Collection.}
Datasets collected through crowd-sourcing or scraping the Internet have the advantage of much better scalability compared to expert-annotated datasets. However, these automatic processes are prone to hidden pitfalls. \citet{gururangan2018annotation,poliak2018hypothesis} show that crowd-sourcing ``Natural Language Inference'' datasets leads to various dataset biases. Similar observations have been made for ``Fact Verification'' datasets~\cite{schuster2019towards}. Splitting data--—for training, testing, and validation—--is another important procedure in creating datasets that can lead to several problems. For example, \citet{geva2019we} show that models may just learn the patterns of certain annotators in a random split. \citet{lewis2020question} demonstrated a significant overlap in current open-domain QA datasets. When present, these unexpected biases or overlaps in datasets can significantly undermine the utility of a dataset and lead to deceptively promising results that are in part due to artifacts of flaws in the dataset rather than successfully modeling the intended task.

\paragraph{Automated Fact Checking for Statements.}
\addtocounter{footnote}{-3}
\stepcounter{footnote}\footnotetext{\url{dataverse.harvard.edu/dataverse/nela}}
\stepcounter{footnote}\footnotetext{\url{github.com/KaiDMML/FakeNewsNet}}
\stepcounter{footnote}\footnotetext{\url{github.com/entitize/Fakeddit}}
Automated fact checking is an important task closely related to unreliable news detection, yet is constructed in a more controlled manner. This task focuses on strictly judging the factuality of one single statement instead of an entire article. \citet{vlachos2014fact} first constructed a dataset with 106 claims from fact-checking websites with paired labels. FEVER~\cite{thorne2018fever} is currently the largest scale fact-verification dataset, where 185,445 claims were generated by modifying sentences from Wikipedia. Both the altered claims and the ground truth supporting evidence are included in the dataset. Existing effective approaches for fact-verification include self-attention based networks~\cite{nie2019combining}, large-scale pretrained transformers~\cite{soleimani2020bert}, neural retrieval methods~\cite{lewis2020retrieval}, and reasoning on semantic-level graphs~\cite{zhong-etal-2020-reasoning}.

%% file: 3collection.tex
\begin{table*}[t]
\centering
\small
\begin{tabular}{llll}
\toprule 
\multicolumn{2}{c}{FakeNewsNet} & \multicolumn{2}{c}{r/Fakeddit} \\
\cmidrule(lr){1-2} \cmidrule(lr){3-4}
Positive Features    & Negative Features   &Positive Features    & Negative Features    \\
\midrule 
season & trump & psbattle & clicks\\
at & brad & says & colorized\\
2018 & pitt & sues & 2018 \\
the & jenner & accused & 2019 \\
awards & jennifier & sells & mrw\\
\bottomrule 
\end{tabular}
\caption{Top five most salient features in the FakeNewsNet dataset and the r/Fakeddit dataset. The features are the highest weighted Bag-of-Word features learned by a Logistic Regression model.}
\label{tab:features}
\end{table*}

\section{Unreliable News Datasets}
Collecting high-quality datasets plays an important role in automatic unreliable news detection research. 
Here we review dataset collection strategies used in constructing recent datasets and point out some hidden pitfalls in these procedures. 

\subsection{Data Collection Strategies}
\label{sec:collect}
Unreliable news detection is usually formalized as a classification task. Accordingly, constructing a dataset requires collecting pairs of news articles and labels. 
\paragraph{News Articles:} Each individual news outlet has its own website where news articles are published. The easiest way to collect a large number of these articles is to simply scrape these websites. Manual annotation or some other mechanism must then be incorporated in order to collect the corresponding labels for each article. Another common way to collect articles is through fact-checking websites. While this approach provides both articles and article-level labels, it normally only provides a limited set of articles. Additionally, scraping these fact-checking websites can lead to additional selection bias in the dataset as highlighted in Section \ref{sec:biases}. 

One other recent trend is collecting posts and corresponding labels from social media~\cite{nakamura2020fakeddit, santia2018buzzface, mitra2015credbank}. While large-scale datasets can be collected through such an approach, they  are often noisier than those collected through traditional news sources, due to a more casual use of language, and a heavier dependency on the context.

\paragraph{News Labels:} The largest challenge in collecting these datasets lies in collecting labels. Manually checking the factuality (or reliability) 
and bias of a single article is time-consuming and requires non-trivial expertise. Modeling such a task through a crowd-sourcing framework is difficult. As such, current research datasets almost exclusively rely on existing resources. As discussed earlier, these resources either provide article-level or site-level labels.  \textbf{Article-level} labels are only available through a few fact-checking websites such as PolitiFact, GossipCop, etc., but the scale is limited since generating these labels is time-consuming and costly.  \textbf{Site-/Outlet-level} labels, on the other hand, available through websites such as MBFC, provide manual labels for each site/outlet. These websites often assign reliable/unreliable or biased/unbiased labels to each news outlet. Many datasets for unreliable news detection assign these site-level labels to all articles in a given site. While these weak or distant labels are not always accurate (one example is shown in Table~\ref{tab:realinfake})
, they provide an easy way to create large-scale datasets. In Table~\ref{tab:datasets}, we highlight three recent large-scale unreliable news datasets along with their data collection procedure.

\begin{table}[t]
\centering
\small
\begin{tabularx}{0.48\textwidth}{l|X}
\toprule
News Outlets & Daily Mail \\
\midrule
Site Label & Unreliable \\
\midrule
Dates & 2018/09/06 \\
\midrule
Title & Roy Moore sues Sasha Baron Cohen \\
\midrule
Article & Failed Senate candidate Roy Moore is suing comedian Sacha Baron Cohen for \$95 million for tricking him into appearing on his Showtime program 'Who is America?'
Moore, whose bid for the Alabama failed in the wake of claims he molested a 14-year-old, filed the lawsuit in Washington DC on Wednesday... \\
\bottomrule
\end{tabularx}
\caption{An example showing a reliable news article from  the ``\textit{Daily Mail}' site which has a ``Low'' factual reporting rate on MBFC. 
Despite coming from a source with low reliability score, the shown article is reliable and very similar to the content on sites with high reliability scores (such as \textit{``BBC''} and \textit{``The Week UK''}) on the same date.}
\label{tab:realinfake}
\end{table}

\subsection{Dataset Selection Biases} \label{sec:biases}
Datasets annotated without expert verification (e.g., through crowdsourcing, automatic web scraping, etc.)
can have some undesired properties that undermine their quality ~\cite{gururangan2018annotation,poliak2018hypothesis,schuster2019towards}. In the following analysis, we choose the FakeNewsNet dataset~\cite{shu2017fake} as a representative example.

We first examine the most salient features in the dataset. To achieve this, we train a Logistic Regression (LR) model on the titles of FakeNewsNet using Bag-of-Words features and show the word features with the highest weights for each class in Table~\ref{tab:features}.\footnote{We also calculated the PMI between the label and  word features as suggested by \citet{gururangan2018annotation} and found the two lists to be very similar. 
} The features in the table show  clear patterns: the top-features for the reliable (positive) class are either stop words (e.g., `at', `the', etc.) 
or words presumably carrying neutral semantics (e.g. `season', `2018', `awards', etc.) while the top features for the unreliable news (negative) class are mostly celebrity names. Using this basic model, we achieve an accuracy of $\sim$78\% 
, while using a BERT-based model that uses both the article and title as input only achieves an incremental improvement yielding an accuracy of 81\% (see Sec. \ref{sec:setup} for detailed model descriptions).
By examining the articles in the dataset, we attribute this to the selection bias exhibited by fact-checking websites. Most unreliable (negative) articles contain  click-bait titles mentioning celebrities, while reliable sources usually have less sensational titles with fewer mentions of celebrities and more diverse keywords. 

\begin{table}[t]
\centering
\small
\begin{tabularx}{0.48\textwidth}{l|X}
\toprule
Label Resource & GossipCop \\
\midrule
Title & NYC terror attack: Celebrities react on social media \\
\midrule

Article & Celebrities are sending their love and support to New York on social media following a terror attack that left eight people dead Tuesday when a truck plowed down pedestrians on a bicycle path near the World Trade Center in Lower Manhattan... \\
\midrule
Label & Unreliable\\
\midrule
News URL & \url{tinyurl.com/yxhvdne6} \\
\bottomrule
\end{tabularx}
\caption{One example from the FakeNewsNet dataset where it is difficult for the article content to support the label. This article contains celebrities' reactions after a terrorist attack. While the article itself does not look like a standard news piece, the reactions in the article are all paired with tweets, so the unreliable label seems to be inconsistent.}
\label{tab:fnn}
\end{table}

Another potential problem is the articles' retrieval framework. FakeNewsNet uses Google search to retrieve the original news article~\cite{shu2017fake}. Internet search engines have proprietary news ranking and verification processes, which means that even when using the original title and source of a given article, the search results might prioritize specific sites over others leading to inaccurate data collection. While \citet{shu2017fake} propose several heuristics to handle these problems, it is unlikely that this noisy process is completely fixed. As a result, we find a few mis-matched title-content pairs where the retrieved article cannot support the label, hence making the example confusing. We show one example with a questionable label in Table~\ref{tab:fnn}, where we suspect the inconsistency is due to the noisy retrieval step.

Finally, the informal nature of user-generated content on social media may be the source of additional biases. In our preliminary experiments, we found that in r/Fakeddit dataset, a simple Bag-of-Words(BoW)-based logistic regression model can reach equal—or even better—performance than the reported BERT-based models (86.91\% vs. 86.44\% in the text-only two-way classification setting), hinting at the strong correlation between the label and lexical inputs.
This is also reflected in the equally confusing most salient features in this dataset shown in Table~\ref{tab:features}.

Since different collection procedures and data resources will lead to different problems, there is no uniform solution to producing a completely bias-free dataset. However, one good test is to check the performance of a simple model such as a BoW-based linear model. 
By analyzing the features learned by the simple model as well as measuring the gap between the performance of a state-of-the-art system and the simple model, one can get a hint of the dataset quality. Unreasonable features, together with small performance gaps, may reveal unwanted biases in the dataset. 
In practice, we also suggest that when developing models using biased datasets to use debiasing techniques (e.g. \citet{schuster2019towards}).

%% file: 4split.tex
\section{Dataset Split Effect}
\label{sec:split}
In this section, we study the effect of time and site/outlet overlap between the training and the evaluation set on the model's performance and show how these confounding factors can impact it.

\subsection{Baseline Models \& Experimental Setup}
\label{sec:setup}
In the following experiments, we use two models: a logistic regression baseline and a state-of-the-art large-scale pretrained Transformer-based model (RoBERTa; ~\citet{liu2019roberta}). 

\paragraph{Logistic Regression (LR):} We use scikit-learn's ~\cite{scikit-learn} implementation of Logistic Regression along with TFIDF-based Bag-of-Words features. We add L2 regularization to the model with a regularization weight of $1.0$ and train the model using L-BFGS. In our experiments, the LR model uses only the title (and not the article body) as the input. 

\paragraph{RoBERTa:} Our implementation is based on the Transformers library~\cite{Wolf2019HuggingFacesTS} and AllenNLP~\cite{Gardner2017AllenNLP}.
We use RoBERTa in two different ways, one takes only the title as the input, the other takes both the title and the article content as the input and formalizes the task as pairwise sentence classification.
Specifically, we concatenate the title and the article content with a \texttt{[SEP]} token in the middle and use different token type embeddings to differentiate between the title and the content. Articles are truncated to fit the 512-token length limit. In the title-only setting, the batch size is set to 32, the learning rate is set to 5e-5, and the model is trained for 3 epochs. In the article+title setting, the batch size is set to 8, the learning rate is set to 2e-5, and the model is trained for 10 epochs. These hyperparameters are set empirically, and our preliminary experiments show that the results are not sensitive to different settings of these hyperparameters.

\begin{table*}[t]
\centering
\small
\begin{tabular}{lcccc}
\toprule 
Model & Input & Random Split    & Source Split (Article) & Source Split (Site)    \\
\midrule 
Majority      & / & 50    & 50        & 69.29 (0.56)    \\
LR & Title      & 77.45    & 67.18 (4.13) & 79.28 (5.27)           \\
% \midrule
RoBERTa & Title & 85.22   & 70.40 (4.28) & 87.83 (10.44) \\
% \midrule
RoBERTa & Title+Article & 96.94 & 80.36 (11.91) & 85.14 (8.00) \\                           
\bottomrule 
\end{tabular}
\caption{Accuracy on validation sets with different split strategies. For ``Source Split', we report the mean and standard deviation (in parentheses) of five different runs. The last column shows the aggregated site-level accuracy.}
\label{tab:split_site}
\end{table*}

\paragraph{Datasets:} Here, our analysis focuses on the 2018 version of the NELA dataset~\cite{horne2018sampling}. Unlike FakeNewsNet, NELA gathers news directly from news outlets, so the influence of selection bias is insignificant. Thus we focus our analysis on other potentially confounding factors in the dataset. We use 
the latest aggregated site-level labels provided in NELA-GT-2019 \cite{gruppi2020nela} and report both the article- and site-level accuracy. For article-level accuracy, we assign the site-level label to all articles from that news outlet and calculate per-article accuracy. For the Source (Site) Split  setting (with no overlap between training and evaluation sites), we also report the site-level accuracy: we aggregate the predictions over individual articles for a given outlet and use the majority prediction as the site-level prediction. We use a balanced label distribution for all dataset splits. 

The results in the third column of Table~\ref{tab:split_site} show the models' performance on the random split, which is the default split method used in most papers, e.g.  ~\cite{nakamura2020fakeddit, horne2018assessing}. 
As the results show, even the simplest logistic regression model achieves an accuracy of over 77\% whereas the RoBERTa model using both title and the news article as the input reaches almost 97\% accuracy. 

\subsection{Effect of Split by Source}
\label{sec:sp_source}
For this experiment, instead of using the standard random split of all the news articles in the dataset, we first randomly split all the sites in the dataset into three disjoint sets (train/dev/test) before adding all articles from each site to their assigned set (train, dev or test). 
We believe this setup is closer to real-world tasks. 
For instance, in order to block all unreliable news sources, one simple—yet useful—approach is to maintain a list of questionable sources. All the news from those sources will be automatically blocked. In this setting, the only remaining task is classifying sources with no or very few annotated examples.
As the results in Table~\ref{tab:split_site} show, there is a significant drop in performance for all the models when compared to the random split. The logistic regression model's performance drops from 77.5 to 67.2\%, and even the more powerful RoBERTa model with both title and article as input drops from 96.9 to 80.4\%, demonstrating the task's significantly increased difficulty. While aggregating article-level results to site-levels can significantly improve the accuracy, we also see a plateauing trend of the performance where adding the article as additional input brings no further improvement to the RoBERTa model. 
Since we subsample the original dataset and balance the number of news articles for each label, the majority baseline (at the article level) is always 50\%. But the site-level majority baseline is well above random (69.29\%). While a new 50\% majority baseline can be achieved by re-subsampling the dataset, the current number also indicates a severe imbalance of dataset size between reliable/unreliable sites which can—potentially—be  exploited by the models.

\begin{table}[t]
\centering
\small
\begin{tabular}{lccc}
\toprule 
Model & Input & Gold Label    & Rand. Label    \\
\midrule 
Majority      & / & 50    & 50          \\
LR & Title      & 77.45    & 66.29           \\
% \midrule
RoBERTa & Title & 85.22   & 74.37  \\
% \midrule
RoBERTa & Title+Article & 96.94 & 95.04  \\                           
\bottomrule 
\end{tabular}
\caption{Article-level accuracy for the random label experiments compared to gold site labels.}
\label{tab:randlabel}
\end{table}

\paragraph{Random Label Experiments:}
For this experiment, we use the original random split strategy. However, we permute all the site-level labels randomly. Hence each label no longer represents the reliability of the site, and is just an arbitrary feature of the site itself. Therefore, the only way for the models to achieve good performance on this task is to memorize the arbitrary site-label mapping. The results in Table~\ref{tab:randlabel} show that the models achieve very high accuracy with the more powerful RoBERTa model with both title and article showing only  \texttildelow2\% accuracy loss when compared to the true labels. These results demonstrate the models' ability to memorize random site-labels, and the similarity between these results and the results on the random splits suggest that the models are bypassing the real task of reliable/unreliable news classification and are just memorizing the site identities.

\paragraph{Performance Variance and Site Similarity Analysis:}
Another interesting observation from the results in Table~\ref{tab:split_site} is that while the performance on every random split is fairly stable, the performance is much more unstable with respect to splitting by source. For example, the RoBERTa (Title+Article) model results have a standard deviation larger than 10 points, with the highest accuracy reaching over $90\%$ and the lowest one below $60\%$.

One potential factor behind the varying performance is the heterogeneity of different news sources (sites). News sites that are similar to those in the training set could be much easier to classify than sites with completely different styles or content.
In this case, even when splitting by site, correlations between the content of similar sites in the training and evaluation sets may drive the generalization performance. To assess this hypothesis, we measure the dependence on the distances between sites in the training and evaluation sets and the model performance at the site level in the evaluation set. Given a set $s$ in the evaluation set, we measure its similarity to all the sites in the training set $t\in S_{train}$. Below we show that higher accuracy on the site $s$ is associated with a higher similarity between $s$ and sites in the training set with the same label $t\in S_{same}$, providing evidence in favor of our hypothesis.

In order to measure the similarity between different sites, we take the representation learned by the RoBERTa model as the representation of the article with a focus on its reliability. Since the RoBERTa model feeds the whole sentence into the multi-layer transformer architecture and feeds the representation of \texttt{[CLS]} token to the downstream classifier~\cite{devlin2019bert,liu2019roberta}, we use the same \texttt{[CLS]} representation as the representation for the whole title+article input.

\begin{table}[t]
\centering
\small
\begin{tabular}{lcc}
\toprule 
Distance & Top 10 Sites & Bottom 10 Sites      \\
\midrule 
l2      & 11.59 & 5.79              \\
cosine & 210.72     & 82.91             \\
% \midrule
MMD & 7.78 & 4.20   \\
% \midrule
CORAL & 29.07 & 14.95  \\                           
\bottomrule 
\end{tabular}
\caption{Average similarity score between sites in the evaluation and training sets.}
\label{tab:sim}
\end{table}

For similarity-metrics between sites, we follow \citet{guo2020multi} and calculate the l2-distance, cosine distance, MMD (maximum mean discrepancy) distance~\cite{gretton2012kernel,li2015generative} and the CORAL (correlation alignment) distance~\cite{sun2016deep,sun2016return}. Following \citet{guo2020multi}, the l2 and cosine distances are calculated by first averaging all the example representations to get the site representation and calculating the distance between site representations; the MMD distance is calculated using an unbiased finite sample estimate from \citet{li2015generative}; and the CORAL distance is calculated by $D_{CORAL}=\frac{1}{4d^2}\lVert C_s-C_t\rVert^2_F$, where $d$ is the feature dimension, $C_s$ and $C_t$ are the co-variance of two sets and $\lVert \cdot \rVert_F^2$ is the squared matrix Frobenius norm. To simplify our analysis, we filter out all the sites containing less than 100 examples (assuming the articles from these sites are too few to significantly influence the model). For every site in the evaluation set $s$, we calculate its distance with respect to every different site $t$ in the training set, and then compare its minimum distance w.r.t the subset of sites with the same gold label $S_{same}$ and the subset of sites with the opposite label $S_{oppo}$, $$\mathit{sim\_score}_s=\frac{\min\limits_{t\in S_{oppo}}\{dist(s,
 t)\}}{\min\limits_{t\in S_{same}}\{dist(s,t)\}}$$

We compute this ratio using all four distances above for the top and bottom 10 sites in the evaluation datasets (ranked based on their accuracy with RoBERTa) and report the mean over all the sites and over all five different random splits in Table~\ref{tab:sim}.
The top 10 sites always have a much larger similarity score than the bottom 10 sites, indicating that they have a much larger similarity with sites in the training sets with the same label. This trend holds across all of the distance metrics. The sensitivity of performance on the site similarity raises additional concerns about how the results in Table~\ref{tab:split_site} may generalize in real-life. As newly emerged unreliable sites are likely to behave differently from old sites, the model's performance may be on the lower end of the variance. 

As a natural extension, we also explored building a model that directly optimizes these site-level distance metrics in order to have better site-level generalization performance. However, in our preliminary results, our model does not show significant improvement from the baseline models. This can also hint at the fact that it is very difficult for these models to extract features that are useful to the task of reliable/unreliable news classification itself and instead learn site-specific features.

\begin{figure}[t]
    \centering
    \includegraphics[width=1.0\linewidth]{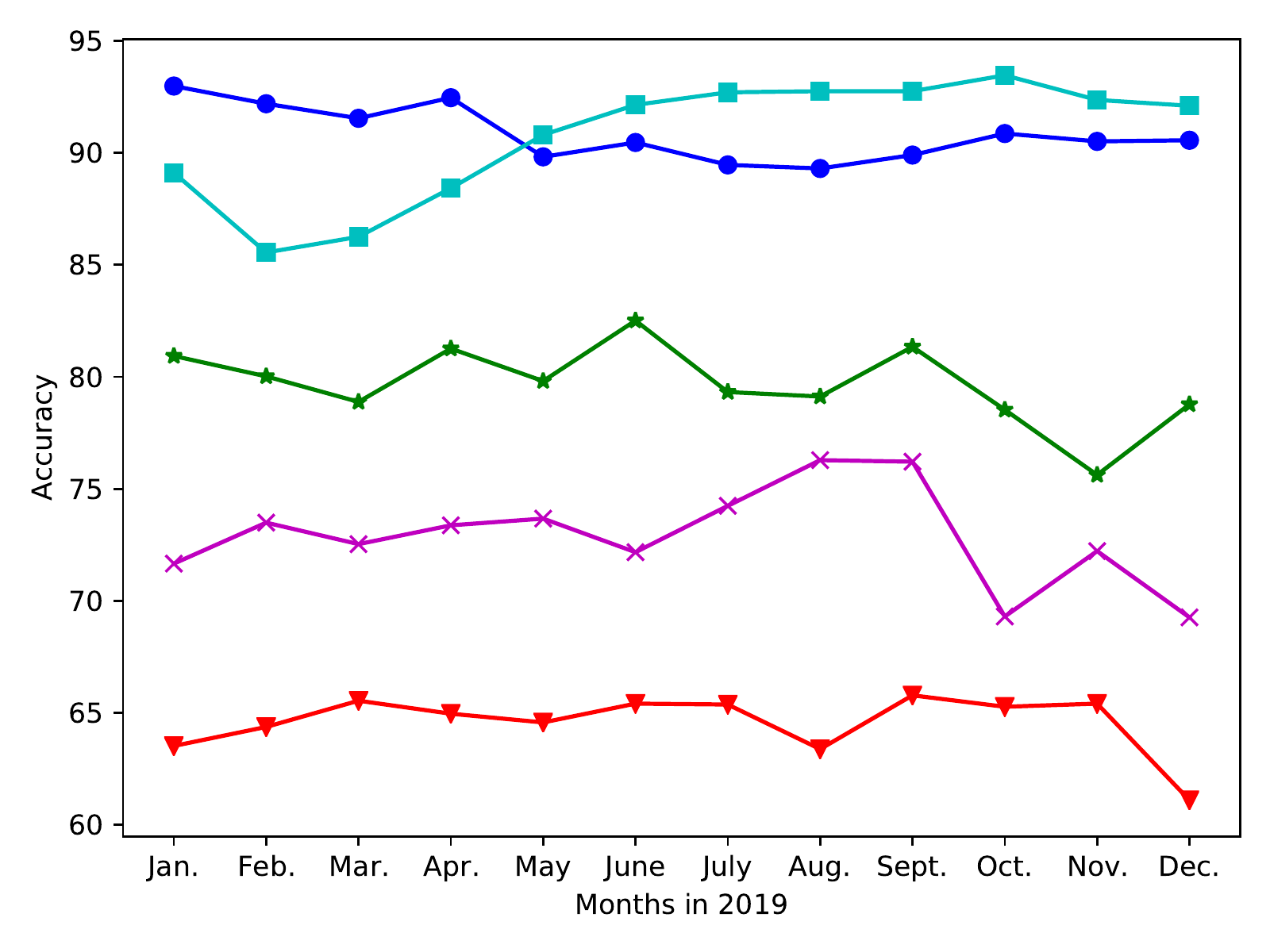}
    \caption{Accuracy of RoBERTa models trained on NELA-GT-2018 and tested on articles from the 12 months covered in the NELA-GT-2019 dataset. The five different lines in the figure represent models trained using five different random site splits.
    }
    % \vspace{-5pt}
    \label{fig:time}
    % \vspace{-5pt}
\end{figure}

\begin{figure*}%
    \centering
    \subfloat[\centering Word cloud of article titles with correct predictions.]{{\includegraphics[width=6.5cm]{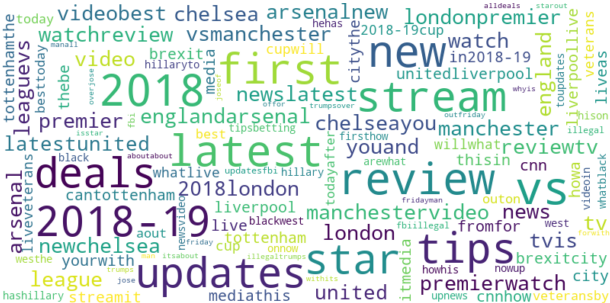} }}%
    \qquad
    \subfloat[\centering Word cloud of article titles with incorrect predictions.]{{\includegraphics[width=6.5cm]{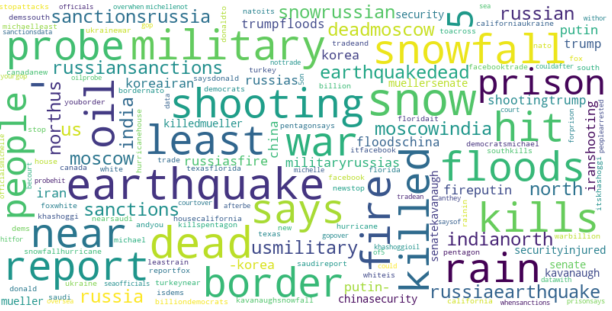} }}%
    \caption{Word cloud of article titles. The words with highest PMI to the prediction correctness of the RoBERTa model are selected.}%
    \label{fig:wc}%
\end{figure*}

\begin{figure*}[t]
    \centering
    \includegraphics[width=0.9\linewidth]{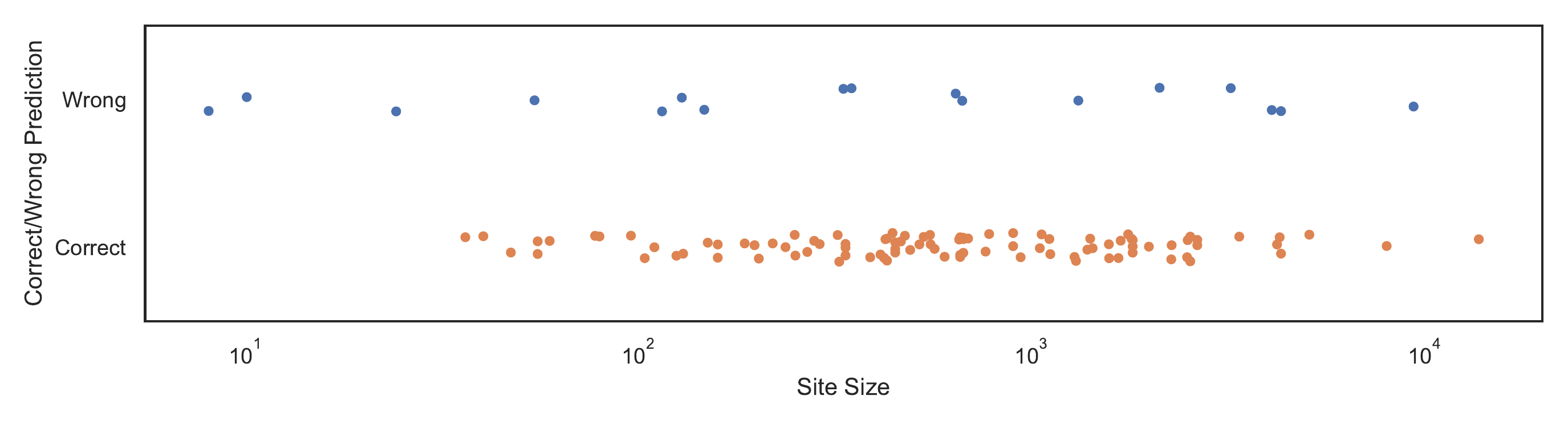}
    \caption{Site-level prediction accuracy of the RoBERTa (Title+Article) model vs. numbers of article in the site (in all five random runs). Blue circles denote wrong predictions and red circles denote correct predictions. 
    }
    % \vspace{-5pt}
    \label{fig:size}
    % \vspace{-5pt}
\end{figure*}

\subsection{Effect of Split By Time}
\label{sec:sp_time}
Another potentially important factor to consider while creating train/test/dev splits for a news-based dataset is time.
As news-worthy events happen everyday, multiple news articles from different outlets can report the same event. For example, in the NELA 2018 dataset~\cite{norregaard2019nela}, within a period of two days (from 2018/10/01 to 2018/10/02), there are more than 100 news articles from over 60 sources about the US-Canada-Mexico trade accord. 
Therefore, by remembering the content of the event from one article, the model can easily predict the label for any related news article. 

To test the effect of time, we examine the model's performance on news articles from a temporally disjoint dataset. Specifically, since all our models are trained on the NELA-GT-2018~\cite{norregaard2019nela}, we use the NELA-GT-2019~\cite{gruppi2020nela} as the evaluation dataset. We split the news articles in 2019 into twelve months and plot the performance trend in Figure~\ref{fig:time}. We can see that, unlike the significant performance drop in the source split experiments, we do not observe a clear correlation between the performance and the length of the time gap. 
Therefore, at least for the current models and datasets, splitting by time does not significantly influence the current results. This finding may result from that the fact that the model is not memorizing the exact events in the training set (this is not limited to the unreliable news domain), or it could be attributed to the noise in the training set (similar events can be reported both in reliable and unreliable sources).
However, we do have to point out that our current observation only holds for the current models, and it is possible for more powerful models to memorize all events. In addition, the widest time gap tested here is still within a couple of years, which is still a relatively short time in terms of news events. A longer time gap (or a major event such as COVID-19) may lead to different behavior by the models. So in practice, we nonetheless suggest splitting datasets by time to avoid these issues.

\subsection{Error Analysis}
\label{sec:error}
Here, we conduct an error analysis to see how the model performs with respect to the variation of some other factors of practical interest, such as topic and site size.
\paragraph{The Influence of Topic in Article-Level Prediction:}
In order to gain better insight on the performance drop in the source split experiments, we perform a deeper investigation of the numbers in Table~\ref{tab:split_site}. We first check whether the models show different performance on different topics. To get a high-level understanding of what the topics are, we look at the titles of articles in the evaluation set and calculate words with the highest PMI with the accuracy of prediction of the RoBERTa model. We then use these PMI values as weights and plot the word cloud figures in Figure~\ref{fig:wc}. In the word cloud of correct predictions, we observe many words related to sports events, while words in the incorrect predictions cloud mostly appear in political news. This is not surprising since there is much more of an incentive to interfere with political news than sports news --- making the need for more robust models even more pressing for real-world applications.

\paragraph{The Influence of Size in Site-Level Prediction:}
Finally, we examine the effect of prediction aggregation from article-level to site-level. Unlike in current datasets where most sites can have hundreds or even thousands of articles, a newly-emerged news outlet waiting for classification may only have a very limited number of articles. Accordingly, while in Table~\ref{tab:split_site}, we see a general improvement of the aggregation, it is also important to check the aggregation effect when the number of articles in a given site is small. 

In Figure~\ref{fig:size} we plot the performance of 5 different runs of the RoBERTa (Title+Article) model against the number of articles on a given site. We can see that the performance is worse when the size of the site is less than 100, demonstrating the difficulty of predicting the reliability of a site given limited resources. It is also surprising to see a significant number of errors even when the site size is over 1000. This indicates the limitation of simply aggregating the site-level prediction at test-time. Capturing the article-site hierarchy in a better way is a potential future research direction.

%% file: 6conclusion.tex
\section{Conclusion}
In this paper, we took a closer look at current large-scale unreliable news detection datasets. We studied their collection procedures and dataset split strategies, and pointed out important flaws in the current approaches. Specifically, we demonstrated that selection bias in dataset collection that often leads to undesired and significant artifacts in these datasets;  highlighting confounding factors (e.g., article source, time) in news datasets that can lead to underestimating the difficulty of the task. Finally we provide suggestions on how to better create and process such datasets in the future. We hope our work leads to more high-quality news datasets and that it inspires further work in this direction.